\documentclass[11pt]{article}

\usepackage{arxiv}
\usepackage{amsmath,amssymb,bm,mathtools}
\usepackage{graphicx}
\usepackage{subcaption}
\usepackage{placeins}
\usepackage{booktabs,array,multirow}
\usepackage{siunitx}
\usepackage{microtype}
\usepackage[numbers,sort&compress]{natbib}
\usepackage[hidelinks]{hyperref}
\usepackage[nameinlink,noabbrev]{cleveref}

\graphicspath{{figures/}}
\newcommand{\R}{\mathbb{R}}
\newcommand{\SO}{\mathrm{SO}}
\newcommand{\SE}{\mathrm{SE}}
\newcommand{\Ad}{\operatorname{Ad}}
\newcommand{\Exp}{\operatorname{exp}}
\renewcommand{\skew}[1]{\left(#1\right)^{\times}}
\newcommand{\I}{\bm{I}}
\newcommand{\zero}{\bm{0}}
\newcommand{\trans}{^{\mathsf T}}
\newsavebox{\maxwidthboxstorage}
\newcommand{\maxwidthbox}[2]{%
  \sbox{\maxwidthboxstorage}{#2}%
  \ifdim\wd\maxwidthboxstorage>#1%
    \resizebox{#1}{!}{\usebox{\maxwidthboxstorage}}%
  \else
    \usebox{\maxwidthboxstorage}%
  \fi}

\title{Accuracy Evaluation of INS/ZUPT Filtering Methods Based on Different Geometric Error Definitions}
\renewcommand{\shorttitle}{Geometric Filtering for INS/ZUPT}
\renewcommand{\headeright}{Preprint}
\renewcommand{\undertitle}{Preprint}

\author{
  Wei Ouyang\textsuperscript{1,*},
  Jiale Han\textsuperscript{2},
  Yarong Luo\textsuperscript{3},
  Maoran Zhu\textsuperscript{2}
  \\[0.5em]
  {\normalfont\small\textsuperscript{1}School of Surveying and Geo-Informatics, Tongji University, Shanghai 200092, China}\\
  {\normalfont\small\textsuperscript{2}School of Automation and Intelligent Sensing, Shanghai Jiao Tong University, Shanghai 200240, China}\\
  {\normalfont\small\textsuperscript{3}School of Robotics, Wuhan University, Wuhan 430072, China}\\
  {\normalfont\small\textsuperscript{*}\texttt{ywoulife@tongji.edu.cn}}
}

\begin{document}
\maketitle

\begin{abstract}
Geometric filters have recently been introduced to improve the accuracy and consistency of inertial-based integrated navigation systems. Error states were defined through specific group operations, introducing state correlations in error definition, which were lacked in the additive error used by a conventional indirect Kalman filter. The desirable consistent filtering models can be obtained based on specific geometric errors. For zero-velocity measurements expressed in the reference frame, this paper derives left-error process and measurement models from invariant filtering, two-frame-group filtering, and equivariant filtering. Importantly, a new group operation is introduced for the left tangent-group equivariant error. The analysis shows that the two-frame-group invariant extended Kalman filter (TFG-IEKF) and the tangent-group equivariant filter (TG-EqF) do not offer a significant consistency advantage over the invariant extended Kalman filter (IEKF). Experiments with an INS/ZUPT measurement system show that, under small initial attitude errors, the conventional indirect extended Kalman filter (EKF) achieves loop-closure position errors below $0.1\%$ of the traveled distance, while the three geometric filters achieve comparable positioning accuracy.
\end{abstract}


\section{Introduction}

In indoor and underground environments where the global navigation satellite system (GNSS) signals are unavailable, inertial navigation systems (INSs) require auxiliary information to suppress unbounded positioning drift. INS/zero-velocity-update (ZUPT) and INS/magnetometer integration are common solutions applied in GNSS-denied environments. The ZUPT is particularly useful because it depends only on the inertial measurement unit (IMU) and is insensitive to external infrastructure \citep{deng2024survey}. Recent studies have considered dynamic step-length constraints for biped inertial positioning \citep{zhao2025biped}, and improved zero-velocity detection and motion constraints were proposed for vehicle navigation \citep{zhang2024vehicle}. The inter-foot ranging was introduced for three-dimensional pedestrian positioning \citep{lu2024interfoot}, and the observability contribution of inter-foot ranging was confirmed in dual-foot systems \citep{zhu2022f2imu}. The non-holonomic constraints widely used in inertial navigation can likewise be interpreted as virtual measurements that set the lateral and vertical body-frame velocities to zero \citep{wang2026pipeline}.

INS/ZUPT systems usually employ an indirect Kalman filter to fuse IMU data with zero-velocity measurements. In recent years, filters based on geometric error definitions have been applied to visual-inertial odometry, lidar-inertial odometry, and simultaneous localization and mapping. Representative methods include the invariant extended Kalman filter (IEKF) and the equivariant filter (EqF), demonstrating excellent convergence and consistency in many inertial sensor-fusion systems. Barrau and Bonnabel established the group-affine condition under which the error dynamics are autonomous, and proposed the IEKF \citep{barrau2017stable,barrau2018invariant} has been applied to numerous inertial-based navigation systems. Hartley et al. applied invariant filtering to legged-robot state estimation and named it as the 'imperfect IEKF' because of the bias-related linearization errors\citep{hartley2020contact}. To further reduce such errors, Barrau and Bonnabel introduced the two-frame group and the TFG-IEKF \citep{barrau2023geometry}, which transfers bias-dependent terms from the navigation-state error dynamics to the bias-error dynamics. Goor et al. developed the general EqF framework \citep{goor2023eqf}, and Fornasier et al. subsequently formulated tangent-group symmetries for biased inertial navigation \citep{fornasier2022biases,fornasier2025symmetries}. However, most existing geometric-filtering studies focus on body-frame observations and right-invariant errors, and only a few works considered the reference-frame measurements, which are naturally paired with left-invariant errors. For instance, Luo investigated the invariant and equivariant Kalman filters for INS/GNSS integrated navigation \citep{luo2023smoother,luo2025lefthanded}. So far the filtering models of the aforementioned left-error geometric filters have not been collective given and their state estimation accuray has not been investigated under the INS/ZUPT system. 

Because INS/ZUPT is a representative system with reference-frame velocity measurements, this paper compares the conventional indirect EKF with three geometric filters based on left-error definitions. Specifically, we derive the process and measurement models of IEKF, TFG-IEKF, and TG-EqF, and their positioning performance is evaluated by using experimental data. In the remaining content of this article, Section 2 directly gives the models of the traditional indirect Kalman filter. Then the left-error filtering models of the invariant EKF, Two-frame group IEKF and tangent group EqF are derived.
Section 3 performs the experiments by using a hand-held data acquisition system. Section 4 conludes this work. Note that the conclusion of this work differs from that reported for many robotic systems with body-frame measurements, such as the invariant/equivariant filtering-based visual-inertial odometry. Discussions are performed for analyzing the conclusion inconsistency for the navigation problems related with reference-frame and body-frame measurements. 

\section{Geometric Error Definitions and Filtering Models}

For a vector $\bm{x}\in\R^3$, the conventional indirect EKF uses the additive error $\delta\bm{x}=\bm{x}-\hat{\bm{x}}$, where $\hat{\bm{x}}$ denotes the estimate. The attitude is represented by a rotation matrix, and its local three-dimensional error is introduced multiplicatively as $\bm{C} \approx \left( \bm{I}_3 - \skew{\delta\bm{\theta}} \right)\hat{\bm{C}},\bm{C} \in \SO\left( 3 \right)$. The traditional EKF error state is
\begin{equation}
\delta \bm{x}_{\mathrm{EKF}}=
\begin{bmatrix}
\delta\bm{\theta}\trans & \delta \bm{v}_{eb}^{e}{\trans} & \delta \bm{r}_{eb}^{e}{\trans} &
\delta \bm{b}_g\trans & \delta \bm{b}_a\trans
\end{bmatrix}\trans ,\qquad \delta \bm{x}_{\mathrm{EKF}} \in \mathbb{R}^{15}
\label{eq:ekf-state}
\end{equation}
in which $\delta \bm{v}_{eb}^{e}$ and $\delta \bm{r}_{eb}^{e}$ are the velocity and position errors, and $\delta \bm{b}_g$ and $\delta \bm{b}_a$ are the gyroscope and accelerometer bias errors.

Using the Earth-fixed frame ($e$-frame) as the reference frame, the INS kinematics are \citep{groves2013principles}
\begin{equation}
\begin{aligned}
\dot{\bm{C}}_b^e &= \bm{C}_b^e\skew{\bm{\omega}_{ib}^b}-\skew{\bm{\omega}_{ie}^e}\bm{C}_b^e,\\
\dot{\bm{v}}_{eb}^e &= \bm{C}_b^e \bm{f}_{ib}^b-2\skew{\bm{\omega}_{ie}^e}\bm{v}_{eb}^e+\bm{g}^e,\\
\dot{\bm{r}}_{eb}^e &= \bm{v}_{eb}^e,\qquad
\dot{\bm{b}}_g=\bm{w}_{bg},\qquad \dot{\bm{b}}_a=\bm{w}_{ba},
\end{aligned}
\label{eq:ins-kinematics}
\end{equation}
where $\bm{\omega}_{ie}^e$ is the Earth rotation rate expressed in the $e$-frame and $\bm{g}^e$ is gravity. 

The IMU measurements satisfy
\begin{equation}
\begin{cases}
\widetilde{\bm{\omega}}_{ib}^b=\bm{\omega}_{ib}^b+\bm{b}_g+\bm{w}_g,\\
\widetilde{\bm{f}}_{ib}^b=\bm{f}_{ib}^b+\bm{b}_a+\bm{w}_a,
\end{cases}
\label{eq:imu-measurements}
\end{equation}
where $\bm{w}_g$ and $\bm{w}_a$ denote the gyroscope and accelerometer noise, respectively.

With the additive errors in \cref{eq:ekf-state}, the linearized error models are \citep{barfoot2017state}
\begin{equation}
\delta\dot{\bm{x}}_{\mathrm{EKF}}=\bm{F}_{\mathrm{EKF}}\delta\bm{x}_{\mathrm{EKF}}+\bm{G}_{\mathrm{EKF}}\bm{w},
\label{eq:ekf-linear}
\end{equation}
with
\begin{equation}
\bm{F}_{\mathrm{EKF}}=
\begin{bmatrix}
-\skew{\bm{\omega}_{ie}^e} & \zero & \zero & \hat{\bm{C}}_b^e & \zero\\
-\skew{\hat{\bm{C}}_b^e\hat{\bm{f}}_{ib}^b} & -2\skew{\bm{\omega}_{ie}^e} & \zero & \zero & \hat{\bm{C}}_b^e\\
\zero & \I_3 & \zero & \zero & \zero\\
\zero & \zero & \zero & \zero & \zero\\
\zero & \zero & \zero & \zero & \zero
\end{bmatrix},
\label{eq:ekf-F}
\end{equation}
\begin{equation}
\bm{G}_{\mathrm{EKF}}=
\begin{bmatrix}
\hat{\bm{C}}_b^e & \zero & \zero & \zero\\
\zero & \hat{\bm{C}}_b^e & \zero & \zero\\
\zero & \zero & \zero & \zero\\
\zero & \zero & \I_3 & \zero\\
\zero & \zero & \zero & \I_3
\end{bmatrix},
\label{eq:ekf-G}
\end{equation}
and
\begin{equation}
\bm{w}=\begin{bmatrix}\bm{w}_g\trans&\bm{w}_a\trans&\bm{w}_{bg}\trans&\bm{w}_{ba}\trans\end{bmatrix}\trans.
\label{eq:ekf-noise}
\end{equation}
Zero velocity is expressed as a velocity observation in the $e$-frame
\begin{equation}
\bm{z}=\bm{v}_{eb}^e+\bm{n}_v,
\label{eq:zupt-measurement}
\end{equation}
whose linearized observation matrix is
\begin{equation}
\bm{H}_{\mathrm{EKF}}=
\begin{bmatrix}\zero&\I_3&\zero&\zero&\zero\end{bmatrix}.
\label{eq:ekf-H}
\end{equation}
The discrete-time indirect EKF is implemented as
\begin{equation}
\begin{aligned}
\hat{\bm{x}}_k^-&=\bm{f}(\hat{\bm{x}}_{k-1}^+),&
\bm{P}_k^-&=\bm{\Phi}_{k,k-1}\bm{P}_{k-1}^+\bm{\Phi}_{k,k-1}\trans+\bm{Q}_{k-1},\\
\bm{K}_k&=\bm{P}_k^-\bm{H}_k\trans(\bm{H}_k\bm{P}_k^-\bm{H}_k\trans+\bm{R}_k)^{-1},&
\delta\hat{\bm{x}}_k&=\bm{K}_k\Delta\bm{z}_k,\\
\bm{P}_k^+&=(\I-\bm{K}_k\bm{H}_k)\bm{P}_k^-,&
\hat{\bm{x}}_k^+&=\hat{\bm{x}}_k^-\boxplus\delta\hat{\bm{x}}_k,
\end{aligned}
\label{eq:ekf-recursion}
\end{equation}
where $\bm{\Phi}_{k,k-1}\simeq\I+\bm{F}_{k-1}T$, $T$ is the INS sampling interval, and
$\bm{w}_{k-1}=\int_{t_{k-1}}^{t_k}\bm{\Phi}_{k,t}\bm{G}(t)\bm{w}(t)\,dt$ defines the discrete process noise. After the measurement update, the nominal state is reset as
\begin{equation}
\begin{aligned}
\bm{C}_b^e&=\Exp\!\left(-\skew{\delta\bm{\theta}}\right)\hat{\bm{C}}_b^e,&
\bm{v}_{eb}^e&=\hat{\bm{v}}_{eb}^e+\delta\bm{v}_{eb}^e,&
\bm{r}_{eb}^e&=\hat{\bm{r}}_{eb}^e+\delta\bm{r}_{eb}^e,\\
\bm{b}_g&=\hat{\bm{b}}_g+\delta\bm{b}_g,&
\bm{b}_a&=\hat{\bm{b}}_a+\delta\bm{b}_a.
\end{aligned}
\label{eq:state-reset}
\end{equation}
The exponential map follows Rodrigues' formula $\Exp(\skew{\bm{\theta}})=\I_3+\sin\theta\,\skew{\bm{n}}+(1-\cos\theta)\bigl(\skew{\bm{n}}\bigr)^2$, with $\bm{\theta}=\theta\bm{n}$.

\subsection{Invariant Extended Kalman Filter}

To account for rigid-body geometry on a manifold, Barrau and Bonnabel extended $\SE(3)$ to the extended-pose group $\SE_2(3)$. For a system $\dot{\bm{\chi}}=f_u(\bm{\chi})$ on a Lie group, the group-affine condition is \citep{barrau2017stable}
\begin{equation}
f_u(\bm{\chi}_1\bm{\chi}_2)=f_u(\bm{\chi}_1)\bm{\chi}_2+\bm{\chi}_1f_u(\bm{\chi}_2)-\bm{\chi}_1f_u(\I)\bm{\chi}_2.
\label{eq:group-affine}
\end{equation}
To satisfy this condition in the Earth-fixed frame, velocity is redefined relative to the inertial frame as $\bar{\bm{v}}^e=\bm{v}_{eb}^e+\skew{\bm{\omega}_{ie}^e}\bm{r}_{eb}^e$. The kinematic models become \citep{wang2025state}
\begin{equation}
\begin{aligned}
\dot{\bm{C}}_b^e&=\bm{C}_b^e\skew{\bm{\omega}_{ib}^b}-\skew{\bm{\omega}_{ie}^e}\bm{C}_b^e,\\
\dot{\bar{\bm{v}}}^{e}&=\bm{C}_b^e\bm{f}_{ib}^b-\skew{\bm{\omega}_{ie}^e}\bar{\bm{v}}^e+\bar{\bm{g}}^e,\\
\dot{\bm{r}}_{eb}^e&=\bar{\bm{v}}^e-\skew{\bm{\omega}_{ie}^e}\bm{r}_{eb}^e,
\end{aligned}
\label{eq:transformed-kinematics}
\end{equation}
where $\bar{\bm{g}}^e=\bm{g}^e+\bigl(\skew{\bm{\omega}_{ie}^e}\bigr)^2\bm{r}_{eb}^e$. Omitting frame subscripts for compactness, and define
\begin{equation}
\bm{\chi}=
\begin{bmatrix}
\bm{C}_b^e&\bar{\bm{v}}^e&\bm{r}_{eb}^e\\
\zero_{1\times3}&1&0\\
\zero_{1\times3}&0&1
\end{bmatrix}\in\SE_2(3).
\label{eq:extended-pose}
\end{equation}
The group-form kinematics are
\begin{equation}
\begin{gathered}
  {\mathbf{\dot {\bm{\chi}} }} = f({\bm{\chi }}) \hfill \\
  \;\;\; = \left[ {\begin{array}{*{20}{c}}
  {{\bm{C}}_b^e{\bm{\omega }}_{ib}^b \times  - {\bm{\omega }}_{ie}^e \times {\bm{C}}_b^e}&{{\bm{C}}_b^e{\bm{f}}_{ib}^b - {\bm{\omega }}_{ie}^e \times {{{\bm{\bar v}}}^e} + {\bm{G}}_{ib}^e}&{{{{\bm{\bar v}}}^e} - {\bm{\omega }}_{ie}^e \times {{\bm{r}}^e}} \\ 
  {{{\bm{0}}_{1 \times 3}}}&0&0 \\ 
  {{{\bm{0}}_{1 \times 3}}}&0&0 
\end{array}} \right] \hfill \\
  \;\;\; = {\bm{\chi }}\left[ {\begin{array}{*{20}{c}}
  {{\bm{\omega }}_{ib}^b \times }&{{\bm{f}}_{ib}^b}&{{{\bm{0}}_{3 \times 1}}} \\ 
  {{{\bm{0}}_{1 \times 3}}}&0&1 \\ 
  {{{\bm{0}}_{1 \times 3}}}&0&0 
\end{array}} \right] + \left[ {\begin{array}{*{20}{c}}
  { - {\bm{\omega }}_{ie}^e \times }&{{\bm{G}}_{ib}^e}&{{{\bm{0}}_{3 \times 1}}} \\ 
  {{{\bm{0}}_{1 \times 3}}}&0&{ - 1} \\ 
  {{{\bm{0}}_{1 \times 3}}}&0&0 
\end{array}} \right]{\bm{\chi }} \hfill \\
  \;\;\; = {\bm{\chi W}} + {\mathbf{U\chi }}. \hfill \\ 
\end{gathered} 
\label{eq:group-kinematics}
\end{equation}

\subsubsection{Left-Invariant Error Definition}

For reference-frame observations, such as GNSS position and velocity, left-invariant errors provide favorable estimation consistency \citep{luo2023smoother,luo2025lefthanded}. The left group error is
\begin{equation}
\bm{\eta}_l=\hat{\bm{\chi}}^{-1}\bm{\chi}=
\begin{bmatrix}
\hat{\bm{C}}_b^{e\trans}\bm{C}_b^e&\hat{\bm{C}}_b^{e\trans}(\bar{\bm{v}}^e-\hat{\bar{\bm{v}}}^{e})&\hat{\bm{C}}_b^{e\trans}(\bm{r}_{eb}^e-\hat{\bm{r}}_{eb}^e)\\
\zero_{1\times3}&1&0\\
\zero_{1\times3}&0&1
\end{bmatrix}.
\label{eq:left-group-error}
\end{equation}
It is related to the Lie algebra by
\begin{equation}
\bm{\eta}_l=\Exp(\bm{\xi}_l^\wedge)=
\begin{bmatrix}
\Exp(\skew{\bm{\xi}_R})&\bm{J}_l(\bm{\xi}_R)\bm{\xi}_v&\bm{J}_l(\bm{\xi}_R)\bm{\xi}_r\\
\zero_{1\times3}&1&0\\
\zero_{1\times3}&0&1
\end{bmatrix},
\label{eq:lie-exp}
\end{equation}

\begin{equation}
{{\bm{J}}_l}\left( {{{\bm{\xi }}_R}} \right) = \frac{{\sin {\xi _R}}}{{{\xi _R}}}{{\bm{I}}_3} + \left( {1 - \frac{{\sin {\xi _R}}}{{{\xi _R}}}} \right){\bm{n}}{{\bm{n}}^T} - \frac{{1 - \cos {\xi _R}}}{{{\xi _R}}}{\skew{\bm{n}}} ,{{\bm{\xi }}_R} = {\xi _R}{\bm{n}}.
\end{equation}
where ${{\bm{\xi }}_l} = {[{{\bm{\xi }}_R}^T,\;{{\bm{\xi }}_v}^T,\;{{\bm{\xi }}_r}^T]^T}$ and $\bm{J}_l(\cdot)$ is the left Jacobian of $\SO(3)$ and
\begin{equation}
\bm{\xi}_l^\wedge=
\begin{bmatrix}
\skew{\bm{\xi}_R}&\bm{\xi}_v&\bm{\xi}_r\\
\zero_{1\times3}&0&0\\
\zero_{1\times3}&0&0
\end{bmatrix}\in\mathfrak{se}_2(3).
\label{eq:lie-algebra}
\end{equation}
For small errors,
\begin{equation}
\hat{\bm{C}}_b^{e\trans}\bm{C}_b^e\approx\I_3+\skew{\bm{\xi}_R},\quad
\bm{\xi}_v\approx\hat{\bm{C}}_b^{e\trans}(\bar{\bm{v}}^e-\hat{\bar{\bm{v}}}^{e}),\quad
\bm{\xi}_r\approx\hat{\bm{C}}_b^{e\trans}(\bm{r}_{eb}^e-\hat{\bm{r}}_{eb}^e).
\label{eq:left-error-components}
\end{equation}
The L-IEKF error state is therefore
\begin{equation}
\delta\bm{x}_{\mathrm{IEKF}}=
\begin{bmatrix}\bm{\xi}_R\trans&\bm{\xi}_v\trans&\bm{\xi}_r\trans&\delta\bm{b}_g\trans&\delta\bm{b}_a\trans\end{bmatrix}\trans.
\label{eq:iekf-state}
\end{equation}

\subsubsection{L-IEKF Filtering Model}

The velocity and position errors in \cref{eq:left-error-components} are coupled to attitude through the group definition. After the linearization, we have the following process model
\begin{equation}
\delta\dot{\bm{x}}_{\mathrm{IEKF}}=\bm{F}_{\mathrm{IEKF}}\delta\bm{x}_{\mathrm{IEKF}}+\bm{G}_{\mathrm{IEKF}}\bm{w},
\label{eq:iekf-linear}
\end{equation}
where
\begin{equation}
\bm{F}_{\mathrm{IEKF}}=
\begin{bmatrix}
-\skew{\hat{\bm{\omega}}_{ib}^b}&\zero&\zero&-\I_3&\zero\\
-\skew{\hat{\bm{f}}_{ib}^b}&-\skew{\hat{\bm{\omega}}_{ib}^b}&\zero&\zero&-\I_3\\
\zero&\I_3&-\skew{\hat{\bm{\omega}}_{ib}^b}&\zero&\zero\\
\zero&\zero&\zero&\zero&\zero\\
\zero&\zero&\zero&\zero&\zero
\end{bmatrix},
\label{eq:iekf-F}
\end{equation}
\begin{equation}
\bm{G}_{\mathrm{IEKF}}=
\begin{bmatrix}
-\I_3&\zero&\zero&\zero\\
\zero&-\I_3&\zero&\zero\\
\zero&\zero&\zero&\zero\\
\zero&\zero&\I_3&\zero\\
\zero&\zero&\zero&\I_3
\end{bmatrix}.
\label{eq:iekf-G}
\end{equation}
Here, $\hat{\bm{\omega}}_{ib}^b=\widetilde{\bm{\omega}}_{ib}^b-\hat{\bm{b}}_g$ and $\hat{\bm{f}}_{ib}^b=\widetilde{\bm{f}}_{ib}^b-\hat{\bm{b}}_a$. The reference-frame velocity innovation can be defined as
\begin{equation}
\begin{aligned}
\Delta\bm{z}
&=\hat{\bm{C}}_b^{e\trans}(\bm{v}^e-\hat{\bm{v}}^e)\\
&=\hat{\bm{C}}_b^{e\trans}(\bar{\bm{v}}^e-\hat{\bar{\bm{v}}}^{e})
-\hat{\bm{C}}_b^{e\trans}\skew{\bm{\omega}_{ie}^e}(\bm{r}^e-\hat{\bm{r}}^e)\\
&\approx \delta\bm{\xi}_v-\skew{\hat{\bm{C}}_b^{e\trans}\bm{\omega}_{ie}^e}\delta\bm{\xi}_r.
\end{aligned}
\label{eq:iekf-innovation}
\end{equation}
Consequently,
\begin{equation}
\bm{H}_{\mathrm{IEKF}}=
\begin{bmatrix}\zero&\I_3&-\skew{\hat{\bm{C}}_b^{e\trans}\bm{\omega}_{ie}^e}&\zero&\zero\end{bmatrix},
\label{eq:iekf-H}
\end{equation}
and the transformed measurement covariance is
\begin{equation}
\bm{R}_k'=\hat{\bm{C}}_b^{e\trans}\bm{R}_k\hat{\bm{C}}_b^e.
\label{eq:iekf-R}
\end{equation}

\subsection{Two-Frame-Group Invariant Extended Kalman Filter}

\subsubsection{Left-Invariant Error Definition}

The two-frame group places the navigation state and IMU biases on a manifold. For $\bm{R}_i\in\SO(3)$, reference-frame vectors $\bm{x}_i\in\R^{3}$, and body-frame vectors $\bm{X}_i\in\R^{3}$, its group operation ‘$\cdot$’ is \citep{barrau2023geometry}
\begin{equation}
(\bm{R}_1,\bm{x}_1,\bm{X}_1)\cdot (\bm{R}_2,\bm{x}_2,\bm{X}_2)
=\left(\bm{R}_1\bm{R}_2,\;\bm{x}_1+\bm{R}_1\bm{x}_2,\;\bm{X}_2+\bm{R}_2\trans \bm{X}_1\right),
\label{eq:tfg-product}
\end{equation}
and the inverse is defined as
\begin{equation}
(\bm{R},\bm{x},\bm{X})^{-1}=\left(\bm{R}^{-1},-\bm{R}^{-1}\bm{x},-\bm{R}\bm{X}\right).
\label{eq:tfg-inverse}
\end{equation}
For attitude, velocity, position, and the two IMU biases, the left group error is
\begin{equation}
\bm{\eta}_l^{\mathrm{TFG}}=\hat{\bm{X}}^{-1}\bm{X}
=\left(\hat{\bm{C}}_b^{e\trans}\bm{C}_b^e,
\hat{\bm{C}}_b^{e\trans}(\bm{x}-\hat{\bm{x}}),
\bm{X}-\bm{C}_b^{e\trans}\hat{\bm{C}}_b^e\hat{\bm{X}}\right).
\label{eq:tfg-error}
\end{equation}
in which, the errors for navigation states are identical to the L-IEKF. Error definitions of two bias errors are
\begin{equation}
\Delta\bm{b}_g=\bm{b}_g-\bm{C}_b^{e\trans}\hat{\bm{C}}_b^e\hat{\bm{b}}_g,\qquad
\Delta\bm{b}_a=\bm{b}_a-\bm{C}_b^{e\trans}\hat{\bm{C}}_b^e\hat{\bm{b}}_a.
\label{eq:tfg-bias-error}
\end{equation}
Using the same navigation-state errors as the IEKF, the TFG-IEKF state is
\begin{equation}
\delta\bm{x}_{\mathrm{TFG}}=
\begin{bmatrix}\bm{\xi}_R\trans&\bm{\xi}_v\trans&\bm{\xi}_r\trans&\Delta\bm{b}_g\trans&\Delta\bm{b}_a\trans\end{bmatrix}\trans.
\label{eq:tfg-state}
\end{equation}

\subsubsection{L-TFG-IEKF Filtering Model}

Based on the error-state definitions in (\ref{eq:tfg-state}), the linearized process model is
\begin{equation}
\delta\dot{\bm{x}}_{\mathrm{TFG}}=\bm{F}_{\mathrm{TFG}}\delta\bm{x}_{\mathrm{TFG}}+\bm{G}_{\mathrm{TFG}}\bm{w},
\label{eq:tfg-linear}
\end{equation}
with
\begin{equation}
\maxwidthbox{0.94\textwidth}{$\displaystyle
\bm{F}_{\mathrm{TFG}}=
\begin{bmatrix}
-\skew{\hat{\bm{\omega}}_{ib}^b+\hat{\bm{b}}_g}&\zero&\zero&-\I_3&\zero\\
-\skew{\hat{\bm{f}}_{ib}^b+\hat{\bm{b}}_a}&-\skew{\hat{\bm{\omega}}_{ib}^b}&\zero&\zero&-\I_3\\
\zero&\I_3&-\skew{\hat{\bm{\omega}}_{ib}^b}&\zero&\zero\\
\skew{\hat{\bm{b}}_g}\skew{\hat{\bm{\omega}}_{ib}^b+\hat{\bm{b}}_g}&\zero&\zero&\skew{\hat{\bm{b}}_g}&\zero\\
\skew{\hat{\bm{b}}_a}\skew{\hat{\bm{\omega}}_{ib}^b+\hat{\bm{b}}_g}&\zero&\zero&\skew{\hat{\bm{b}}_a}&\zero
\end{bmatrix}.$}
\label{eq:tfg-F}
\end{equation}
\begin{equation}
\bm{G}_{\mathrm{TFG}}=
\begin{bmatrix}
-\I_3&\zero&\zero&\zero\\
\zero&-\I_3&\zero&\zero\\
\zero&\zero&\zero&\zero\\
\skew{\hat{\bm{b}}_g}&\zero&\I_3&\zero\\
\skew{\hat{\bm{b}}_a}&\zero&\zero&\I_3
\end{bmatrix}.
\label{eq:tfg-G}
\end{equation}
In \cref{eq:tfg-F,eq:tfg-G}, $\hat{\bm{\omega}}_{ib}^b+\hat{\bm{b}}_g=\widetilde{\bm{\omega}}_{ib}^b$ and $\hat{\bm{f}}_{ib}^b+\hat{\bm{b}}_a=\widetilde{\bm{f}}_{ib}^b$ are the uncorrected IMU measurements. Because the TFG-IEKF changes only the bias-error definition relative to the IEKF, it uses the same observation model as \cref{eq:iekf-H}.

\subsection{Tangent-Group Equivariant Filter}

The EqF based on the tangent group represents body-frame vectors in the Lie-algebra space \citep{fornasier2025symmetries}. To make the inertial system equivariant, a virtual body-frame velocity $\widetilde{\bm{v}}$ and its bias $\bm{b}_v$ are introduced into the position differential equation as
\begin{equation}
\dot{\bm{r}}^e=\bm{C}_b^e(\widetilde{\bm{v}}-\bm{b}_v-\bm{w}_v)+\bar{\bm{v}}^e-\skew{\bm{\omega}_{ie}^e}\bm{r}^e.
\label{eq:virtual-velocity}
\end{equation}
The state manifold contains the extended pose, IMU biases, and virtual-velocity bias. The EqF with the tangent group defines the manifold $\mathcal{M}: = \mathcal{S}{\mathcal{E}_2}(3) \times {\mathbb{R}^9}$, and the elements on the manifold is given by $\xi :({\mathbf{T}},{\mathbf{b}}) \in \mathcal{M}$, ${\mathbf{T}} = \left( {{\mathbf{R}},{\mathbf{v}},{\mathbf{r}}} \right) \in \mathcal{S}{\mathcal{E}_2}(3)$, ${\mathbf{b}} = \left( {{{\mathbf{b}}_g},{{\mathbf{b}}_a},{{\mathbf{b}}_v}} \right) \in {\mathbb{R}^9}$. The velocity bias satisfies the model ${{\mathbf{\dot b}}_v} = {{\mathbf{w}}_{bv}}$. Let $X=(C_X,\gamma_X)$ and $Y=(C_Y,\gamma_Y)$ be elements of the semidirect product group $\mathbf{G}=\SE_2(3)\ltimes\mathfrak{se}_2(3)$, in which $C = (R,v,r) \in S{E_2}(3)$, $\gamma  = {\left( {{\gamma _g},{\gamma _a},{\gamma _v}} \right)^ \wedge } \in \mathfrak{s}{\mathfrak{e}_2}(3)$. The tangent-group product and inverse are defined as
\begin{equation}
XY=\left(C_XC_Y,\gamma_X+\Ad_{C_X}\gamma_{Y}\right),
\label{eq:tg-right-product}
\end{equation}
\begin{equation}
X^{-1}=\left(C_X^{-1},-\Ad_{C_X^{-1}}\gamma_X\right).
\label{eq:tg-right-inverse}
\end{equation}
For $C=(R,v,r)\in\SE_2(3)$, the adjoint matrices are
\begin{equation}
\Ad_{C}=
\begin{bmatrix}
\bm{R}&\zero&\zero\\
\skew{\bm{v}}\bm{R}&\bm{R}&\zero\\
\skew{\bm{r}}\bm{R}&\zero&\bm{R}
\end{bmatrix},\qquad
\Ad_{\bm{C}^{-1}}=
\begin{bmatrix}
\bm{R}\trans&\zero&\zero\\
-\bm{R}\trans\skew{\bm{v}}&\bm{R}\trans&\zero\\
-\bm{R}\trans\skew{\bm{r}}&\zero&\bm{R}\trans
\end{bmatrix}.
\label{eq:adjoint}
\end{equation}
Note that the authors of EqF adopted the bold font and slim font to denote the members on manifold and group\cite{fornasier2025symmetries,fornasier2022biases}, respectively. In (\ref{eq:adjoint}), the adjiont matrix of the group member is denoted with bold-font matrices and vectors as usual.

The EqF uses the right action of group $X = \left( {C,\gamma } \right)$ on the manifold $\xi  = ({\mathbf{T}},{\mathbf{b}})$ 
\begin{equation}
\phi(X,\xi)=\left(\bm{T}C,\Ad_{C^{-1}}^{\vee}(\bm{b}-\gamma^{\vee})\right)\in\mathcal M,
\label{eq:right-action}
\end{equation}
which maps a group estimate $\hat X = \left( {\hat C,\hat \gamma } \right)$ to the state manifold as
\begin{equation}
\hat{\xi}=\phi(\hat{X},\xi_I)
=\left(\hat{\bm{C}},\Ad_{\hat{C}^{-1}}^{\vee}(-\hat{\bm{\gamma}}^{\vee})\right)
=\left(\hat{\bm{T}},\hat{\bm{b}}\right).
\label{eq:right-estimate}
\end{equation}
and therefore ${\bm{\hat b}} = A{d_{{{\hat C}^{ - 1}}}}\left( { - {{\hat \gamma }^ \vee }} \right)$.

The corresponding right error as
\begin{equation}
\begin{gathered}
  e: = \phi \left( {{{\hat X}^{ - 1}},\xi } \right) \hfill \\
  \;\;\; = \left( {{\mathbf{T}}{{\hat C}^{ - 1}},Ad_{\hat C}^ \vee \left( {{\mathbf{b}} + A{d_{{{\hat C}^{ - 1}}}}{{\hat \gamma }^ \vee }} \right)} \right) \hfill \\
  \;\;\; = \left( {{\mathbf{T}}{{\hat C}^{ - 1}},Ad_{\hat C}^ \vee {\mathbf{b}} + {{\hat \gamma }^ \vee }} \right) \hfill \\
  \;\;\; = \left( {{\mathbf{T}}{{{\mathbf{\hat T}}}^{ - 1}},Ad_{{\mathbf{\hat T}}}^ \vee ({\mathbf{b}} - {\mathbf{\hat b}})} \right) \hfill \\ 
\end{gathered} 
\label{eq:right-error}
\end{equation}
This right error is suitable for the information fusion system receiving body-frame observations, such as visual, lidar, and odometer measurements, but not for the reference-frame velocity measurement considered here. As a result, another group operation and error function are further proposed in the sequel.

\subsubsection{Left Group Error for the TG-EqF}

For reference-frame measurements, however, we define a compatible left group product and inverse as
\begin{equation}
{X}{Y}=\left({C}_X{C}_Y,{\gamma}_Y+\Ad_{{C}_Y^{-1}}{\gamma}_X\right),\qquad
{X}^{-1}=\left({C}_X^{-1},-\Ad_{{C}_X}{\gamma}_X\right).
\label{eq:tg-left-product}
\end{equation}
The associated left group action on the state manifold is further defined as
\begin{equation}
\varphi({X},{\xi})=\left({C}{\bm{T}},\Ad_{{\bm{T}}^{-1}}^{\vee}{\gamma}^{\vee}+{\bm{b}}\right),
\label{eq:left-action}
\end{equation}
and the estimated state on manifold can be obtained from the identity origin $\bm{\xi}_I$ as
\begin{equation}
\varphi(\hat{{X}},{\xi}_I)=\left(\hat{{C}},\hat{{\gamma}}^{\vee}\right)=\left(\hat{\bm{T}},\hat{\bm{b}}\right).
\label{eq:left-estimate}
\end{equation}
The resulting left error on the manifold can be defined as
\begin{equation}
\begin{gathered}
  e = \varphi \left( {{{\hat X}^{ - 1}},\xi } \right) \hfill \\
  \;\; = \left( {{{\hat C}^{ - 1}}{\bm{T}},Ad_{{{\bm{T}}^{ - 1}}}^ \vee \left( { - A{d_{\hat C}}{{\hat \gamma }^ \vee }} \right) + {\bm{b}}} \right) \hfill \\
  \;\; = \left( {{{\hat C}^{ - 1}}{\bm{T}},{\bm{b}} - Ad_{{{\bm{T}}^{ - 1}}}^ \vee A{d_{\hat C}}{{\hat \gamma }^ \vee }} \right) \hfill \\
  \;\; = \left( {{{{\bm{\hat T}}}^{ - 1}}{\bm{T}},{\bm{b}} - Ad_{{{\bm{T}}^{ - 1}}{\bm{\hat T}}}^ \vee {\bm{\hat b}}} \right) \hfill \\ 
\end{gathered}.
\label{eq:left-tg-error}
\end{equation}
In contrast with the left invariant error, the errors for body-frame vectors are computed as
\begin{equation}
\maxwidthbox{0.96\textwidth}{$\displaystyle
\bm{b}-\Ad_{\bm{T}^{-1}\hat{\bm{T}}}^{\vee}\hat{\bm{b}}=
\begin{bmatrix}\bm{b}_g\\\bm{b}_a\\\bm{b}_v\end{bmatrix}
-
\begin{bmatrix}
\bm{C}_b^{e\trans}\hat{\bm{C}}_b^e&\zero&\zero\\
\skew{\bm{C}_b^{e\trans}(\hat{\bar{\bm{v}}}^e-\bar{\bm{v}}^e)}\bm{C}_b^{e\trans}\hat{\bm{C}}_b^e&\bm{C}_b^{e\trans}\hat{\bm{C}}_b^e&\zero\\
\skew{\bm{C}_b^{e\trans}(\hat{\bm{r}}^e-\bm{r}^e)}\bm{C}_b^{e\trans}\hat{\bm{C}}_b^e&\zero&\bm{C}_b^{e\trans}\hat{\bm{C}}_b^e
\end{bmatrix}
\begin{bmatrix}\hat{\bm{b}}_g\\\hat{\bm{b}}_a\\\hat{\bm{b}}_v\end{bmatrix}.$}
\label{eq:tg-bias-vector}
\end{equation}
The first-order approximations are further denoted by
\begin{equation}
\Delta\bm{b}_g=\bm{b}_g-\bm{C}_b^{e\trans}\hat{\bm{C}}_b^e\hat{\bm{b}}_g
\approx \bm{b}_g-\hat{\bm{b}}_g-\skew{\hat{\bm{b}}_g}\bm{\xi}_R,
\label{eq:tg-bg}
\end{equation}
\begin{equation}
\begin{aligned}
\Delta\bm{b}_a
&=\bm{b}_a-\skew{\hat{\bm{C}}_b^{e\trans}(\hat{\bar{\bm{v}}}^e-\bar{\bm{v}}^e)}\bm{C}_b^{e\trans}\hat{\bm{C}}_b^e\hat{\bm{b}}_g
-\bm{C}_b^{e\trans}\hat{\bm{C}}_b^e\hat{\bm{b}}_a\\
&\approx \bm{b}_a+\skew{\hat{\bm{C}}_b^{e\trans}(\hat{\bar{\bm{v}}}^e-\bar{\bm{v}}^e)}\hat{\bm{b}}_g
-\bm{C}_b^{e\trans}\hat{\bm{C}}_b^e\hat{\bm{b}}_a,
\end{aligned}
\label{eq:tg-ba}
\end{equation}
and
\begin{equation}
\begin{aligned}
\Delta\bm{b}_v
&=\bm{b}_v-\skew{\hat{\bm{C}}_b^{e\trans}(\hat{\bm{r}}^e-\bm{r}^e)}\bm{C}_b^{e\trans}\hat{\bm{C}}_b^e\hat{\bm{b}}_g
-\bm{C}_b^{e\trans}\hat{\bm{C}}_b^e\hat{\bm{b}}_v\\
&\approx \bm{b}_v+\skew{\hat{\bm{C}}_b^{e\trans}(\bm{r}^e-\hat{\bm{r}}^e)}\hat{\bm{b}}_g
-\bm{C}_b^{e\trans}\hat{\bm{C}}_b^e\hat{\bm{b}}_v.
\end{aligned}
\label{eq:tg-bv}
\end{equation}
Then, the TG-EqF error state is
\begin{equation}
\delta\bm{x}_{\mathrm{TG}}=
\begin{bmatrix}
\bm{\xi}_R\trans&\bm{\xi}_v\trans&\bm{\xi}_r\trans&\Delta\bm{b}_g\trans&\Delta\bm{b}_a\trans&\Delta\bm{b}_v\trans
\end{bmatrix}\trans.
\label{eq:tg-state}
\end{equation}

\subsubsection{L-TG-EqF Filtering Model}

Based on the error definitions of IMU and virtual velocity biases, the linearized process model is computed as
\begin{equation}
\delta\dot{\bm{x}}_{\mathrm{TG}}=\bm{F}_{\mathrm{TG}}\delta\bm{x}_{\mathrm{TG}}+\bm{G}_{\mathrm{TG}}\bm{w},
\label{eq:tg-linear}
\end{equation}
where
\begin{equation}
\bm{w}=\begin{bmatrix}\bm{w}_g\trans&\bm{w}_a\trans&\bm{w}_v\trans&\bm{w}_{bg}\trans&\bm{w}_{ba}\trans &\bm{w}_{bv}\trans\end{bmatrix}\trans,
\label{eq:tg-noise}
\end{equation}

\begin{equation}
\resizebox{0.99\textwidth}{!}{$
\bm{F}_{\mathrm{TG}}=
\begin{bmatrix}
-\skew{\hat{\bm{\omega}}+\hat{\bm{b}}_g}&\zero&\zero&-\I_3&\zero&\zero\\
-\skew{\hat{\bm{f}}+\hat{\bm{b}}_a}&-\skew{\hat{\bm{\omega}}+\hat{\bm{b}}_g}&\zero&\zero&-\I_3&\zero\\
\zero&\I_3&-\skew{\hat{\bm{\omega}}+\hat{\bm{b}}_g}&\zero&\zero&-\I_3\\
\skew{\hat{\bm{b}}_g}\skew{\hat{\bm{\omega}}+\hat{\bm{b}}_g}&\zero&\zero&\skew{\hat{\bm{b}}_g}&\zero&\zero\\
\skew{\hat{\bm{b}}_a}\skew{\hat{\bm{\omega}}+\hat{\bm{b}}_g}+\skew{\hat{\bm{b}}_g}\skew{\hat{\bm{f}}+\hat{\bm{b}}_a}
&\skew{\hat{\bm{b}}_g}\skew{\hat{\bm{\omega}}+\hat{\bm{b}}_g}&\zero&\skew{\hat{\bm{b}}_a}&\skew{\hat{\bm{b}}_g}&\zero\\
\skew{\hat{\bm{b}}_v}\skew{\hat{\bm{\omega}}+\hat{\bm{b}}_g}&-\skew{\hat{\bm{b}}_g}&\skew{\hat{\bm{b}}_g}\skew{\hat{\bm{\omega}}+\hat{\bm{b}}_g}&\skew{\hat{\bm{b}}_v}&\zero&\skew{\hat{\bm{b}}_g}
\end{bmatrix}.$}
\label{eq:tg-F}
\end{equation}

\begin{equation}
\bm{G}_{\mathrm{TG}}=
\begin{bmatrix}
-\I_3&\zero&\zero&\zero&\zero&\zero\\
\zero&-\I_3&\zero&\zero&\zero&\zero\\
\zero&\zero&-\I_3&\zero&\zero&\zero\\
\skew{\hat{\bm{b}}_g}&\zero&\zero&\I_3&\zero&\zero\\
\skew{\hat{\bm{b}}_a}&\skew{\hat{\bm{b}}_g}&\zero&\zero&\I_3&\zero\\
\skew{\hat{\bm{b}}_v}&\zero&\skew{\hat{\bm{b}}_g}&\zero&\zero&\I_3
\end{bmatrix}.
\label{eq:tg-G}
\end{equation}
Because the EqF includes the virtual velocity bias, its observation matrix extends \cref{eq:iekf-H} to
\begin{equation}
\bm{H}_{\mathrm{TG}}=
\begin{bmatrix}\zero&\I_3&-\skew{\hat{\bm{C}}_b^{e\trans}\bm{\omega}_{ie}^e}&\zero&\zero&\zero\end{bmatrix}.
\label{eq:tg-H}
\end{equation}

\section{Experiments and Analysis}

\subsection{INS/ZUPT Positioning System}

\begin{figure}[t]
  \centering
  \begin{subfigure}[b]{0.48\textwidth}
    \centering
    \includegraphics[width=4cm,height=5cm,keepaspectratio]{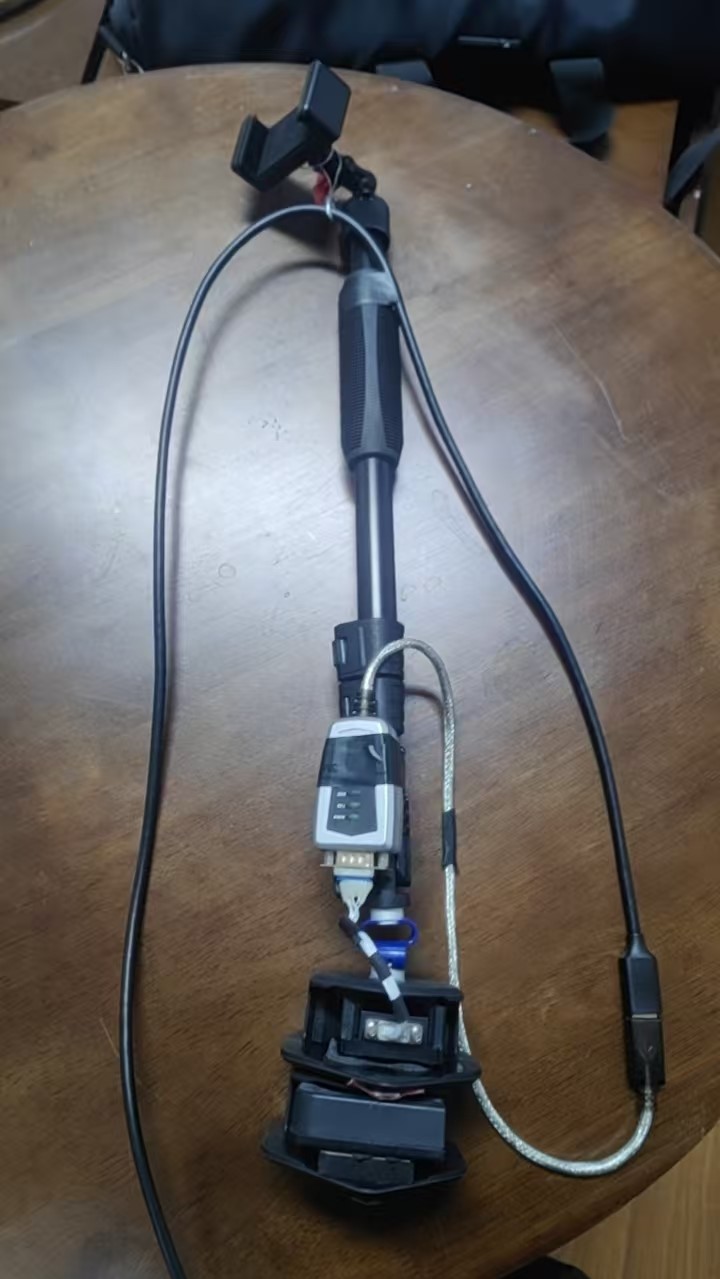}
    \caption{Hand-held INS/ZUPT data-acquisition system.}
  \end{subfigure}\hfill
  \begin{subfigure}[b]{0.48\textwidth}
    \centering
    \includegraphics[width=\linewidth,height=5cm,keepaspectratio]{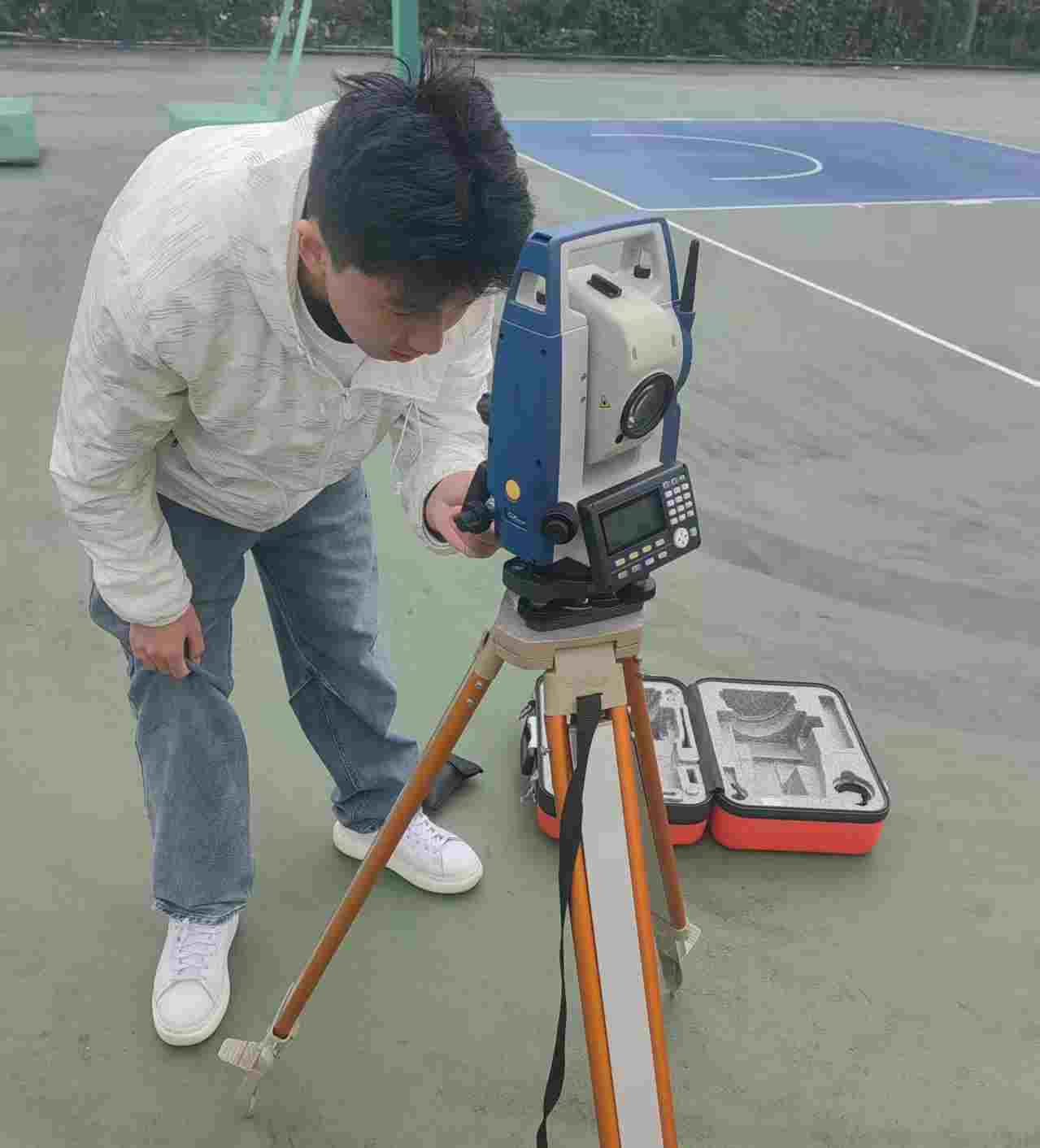}
    \caption{Total-station measurements of reference points.}
  \end{subfigure}
  \caption{Data-acquisition system and total-station point measurements.}
  \label{fig:system}
\end{figure}

The three geometric filters were evaluated in an INS/ZUPT system comprising a tri-axis gyroscope and a tri-axis accelerometer. The data-acquisition device is shown in \cref{fig:system}. Unlike the conventional foot-mounted pedestrian navigation system, the hand-held measurement system was operated in a walking-and-stationary positioning pattern. The ground-truth coordinates of control points along the indoor trajectory were measured with a total station. The specifications of MEMS IMU  are listed in \cref{tab:imu}.

Because the study concerns relative positioning in GNSS-denied environments, multiple closed-loop trajectories were collected by making the physical start and end points coincide. Loop-closure position error was used as the primary metric. The conventional indirect EKF served as the baseline. For every method, the initial attitude matrix was set to the identity matrix and the standard deviation of the zero-velocity measurement was set to \SI{0.01}{\meter\per\second}.

\begin{table}[t]
\centering
\caption{Specifications of the MEMS IMU.}
\label{tab:imu}
\begin{tabular}{lcc}
\toprule
Specification & Gyroscope & Accelerometer\\
\midrule
Noise random walk & $0.015~\mathrm{deg}/\sqrt{\mathrm{h}}$ & $0.017~(\mathrm{m}/\mathrm{s})/\sqrt{\mathrm{h}}$\\
Bias instability & $0.5~\mathrm{deg}/\mathrm{h}$ & $10~\mu g$\\
Measurement range & $500~\mathrm{deg}/\mathrm{s}$ & $8g$\\
\bottomrule
\end{tabular}
\end{table}

\begin{figure}[t]
  \centering
  \begin{subfigure}[b]{0.48\textwidth}
    \centering
    \includegraphics[width=\linewidth,height=5.2cm,keepaspectratio]{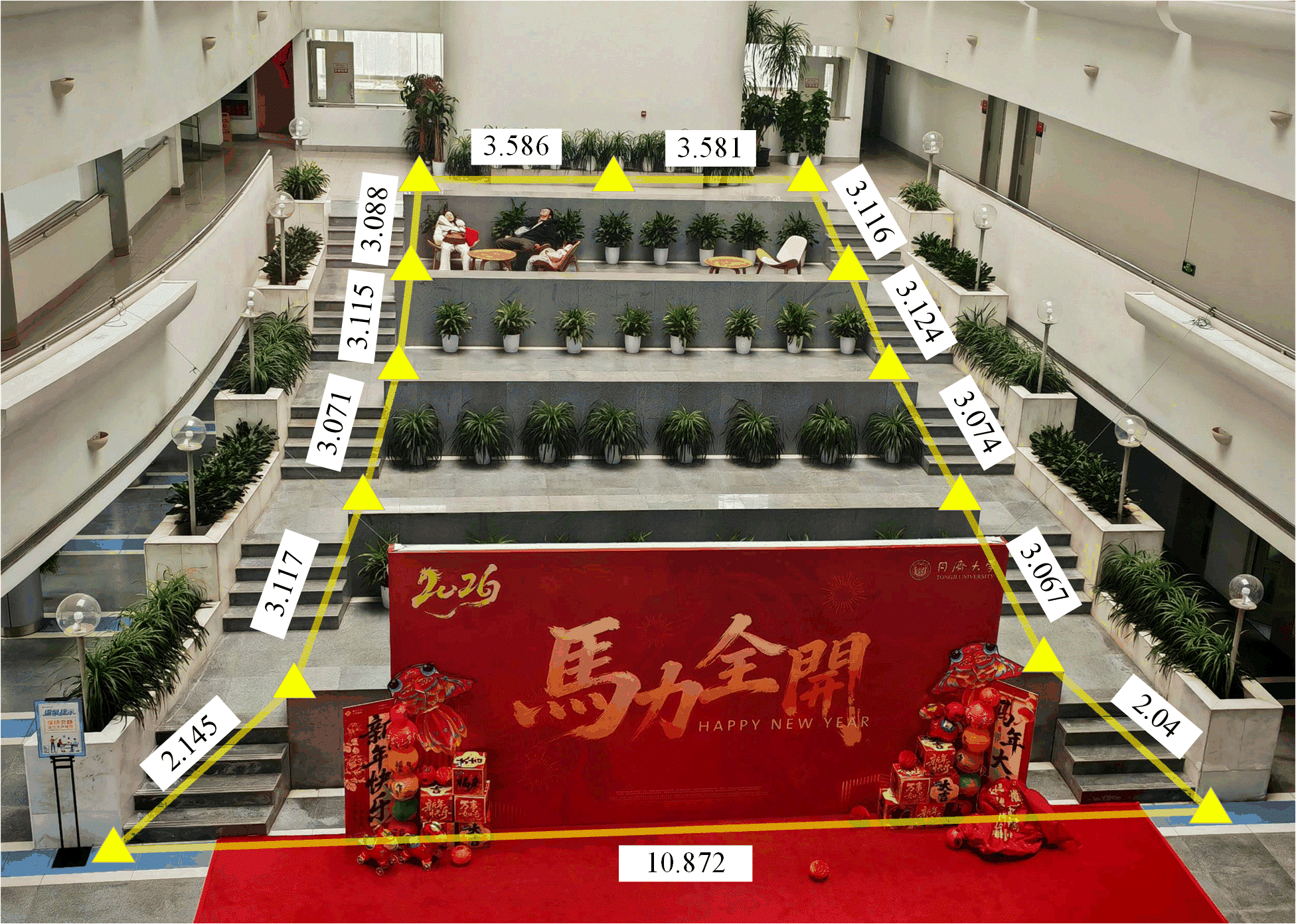}
    \caption{Indoor test scenario.}
  \end{subfigure}\hfill
  \begin{subfigure}[b]{0.48\textwidth}
    \centering
    \includegraphics[width=\linewidth,height=5.2cm,keepaspectratio]{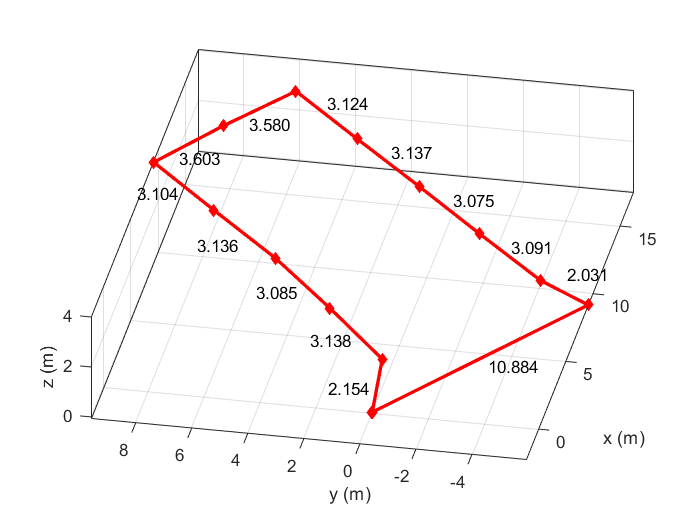}
    \caption{Estimated stationary-point positions.}
  \end{subfigure}
  \caption{Indoor experiment and positioning result.}
  \label{fig:indoor}
\end{figure}

The system was first used to survey the stepped architectural structure of Rui'an Building at Tongji University. The trajectory began at the lower-left corner of the stairs and returned to the same point. Stationary measurement points were placed at both ends of each landing, and their coordinates were measured by the total station. As shown in \cref{fig:indoor}, the indirect EKF accurately recovered the structure. The total line length was approximately \SI{47.14}{\meter}, the loop-closure error was about \SI{0.02}{\meter}, and the relative closure error was below $0.05\%D$, where $D$ denotes the length of the total line distance. The lengths of each line segment are shown in \cref{tab:indoor-distance}, which confirm centimeter-level accuracy over this small indoor scenario.

\begin{table}[t]
\centering
\caption{Errors in distances between adjacent indoor control points (m).}
\label{tab:indoor-distance}
\setlength{\tabcolsep}{3.0pt}
\resizebox{\textwidth}{!}{%
\begin{tabular}{lrrrrrrrrrrrrr}
\toprule
Segment & 1 & 2 & 3 & 4 & 5 & 6 & 7 & 8 & 9 & 10 & 11 & 12 & 13\\
\midrule
Reference & 2.145 & 3.117 & 3.071 & 3.115 & 3.088 & 3.586 & 3.581 & 3.116 & 3.124 & 3.074 & 3.067 & 2.040 & 10.872\\
Measured  & 2.154 & 3.138 & 3.085 & 3.136 & 3.104 & 3.603 & 3.580 & 3.124 & 3.137 & 3.075 & 3.091 & 2.031 & 10.884\\
Absolute error & 0.009 & 0.021 & 0.014 & 0.021 & 0.016 & 0.017 & 0.001 & 0.008 & 0.013 & 0.001 & 0.024 & 0.009 & 0.009\\
\bottomrule
\end{tabular}}
\end{table}

\subsection{Comparative Experiments on Geometric Filters}

\begin{figure}[t!]
  \centering
  \begin{subfigure}[b]{0.32\textwidth}\includegraphics[width=\linewidth]{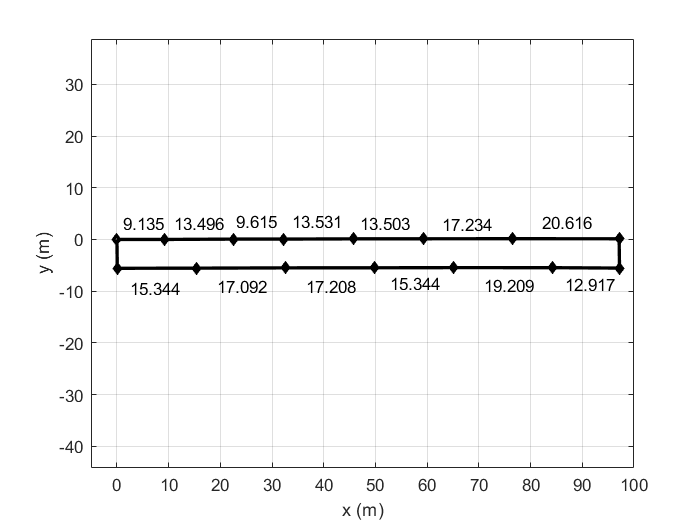}\caption{Data 1}\end{subfigure}
  \begin{subfigure}[b]{0.32\textwidth}\includegraphics[width=\linewidth]{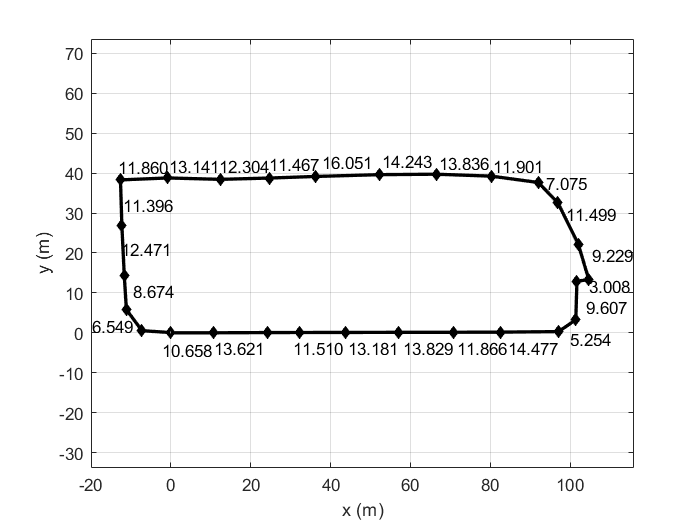}\caption{Data 2}\end{subfigure}
  \begin{subfigure}[b]{0.32\textwidth}\includegraphics[width=\linewidth]{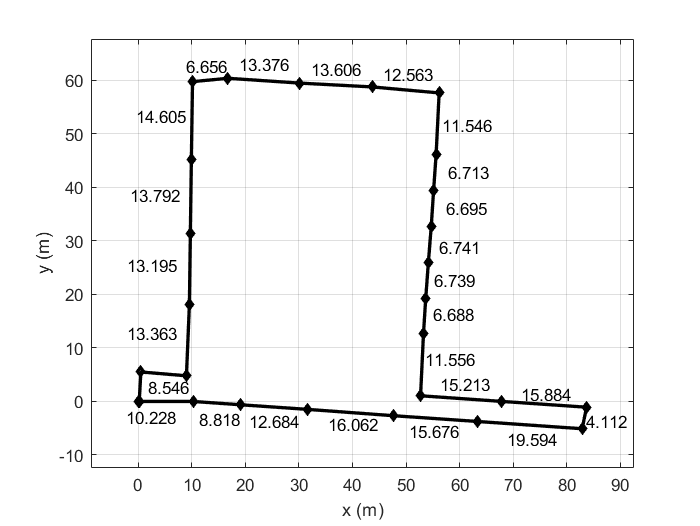}\caption{Data 3}\end{subfigure}
  \begin{subfigure}[b]{0.32\textwidth}\includegraphics[width=\linewidth]{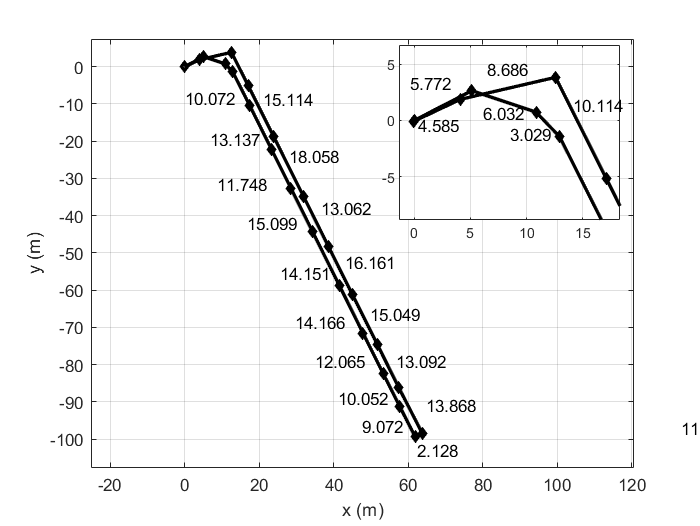}\caption{Data 4}\end{subfigure}
  \begin{subfigure}[b]{0.32\textwidth}\includegraphics[width=\linewidth]{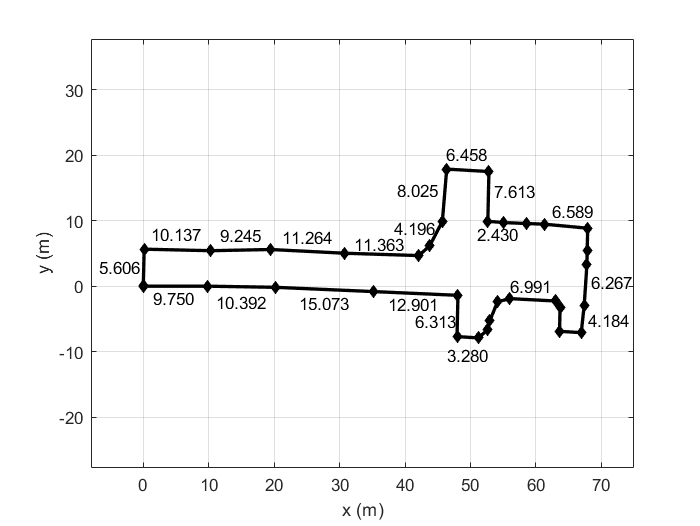}\caption{Data 5}\end{subfigure}
  \begin{subfigure}[b]{0.32\textwidth}\includegraphics[width=\linewidth]{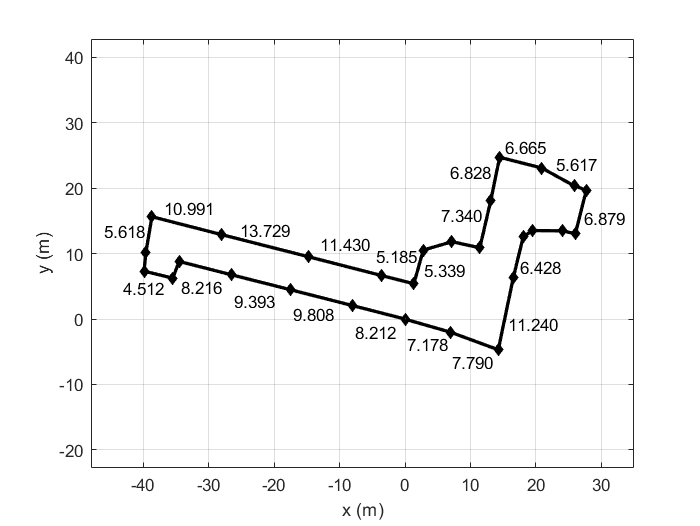}\caption{Data 6}\end{subfigure}
  \caption{Positioning results for the six outdoor experiments.}
  \label{fig:outdoor}
\end{figure}

Six outdoor data were collected to compare the three geometric filters with the conventional indirect EKF. The trajectories in \cref{fig:outdoor} were \SIrange{180}{300}{\meter} long with walking and occasional stationary ground contact, and each stationary interval lasting approximately \SI{2}{\second}. The initial outdoor heading was coarsely determined from the Magnetometer, while roll and pitch were initialized from accelerometer leveling.

The loop-closure results in \cref{tab:closure} show that the differences among the mean errors of the four methods are below \SI{0.01}{\meter}, and the relative error is below $0.1\%D$. Considering unavoidable operational offsets when physically closing the loop, the four methods can be regarded as having essentially the same positioning accuracy. Dataset statistics are summarized in \cref{tab:dataset}. Data 2 and Data 3 are the longest trajectories and exhibit correspondingly larger accumulated errors.

\begin{table}[t]
\centering
\caption{Loop-closure position errors for the outdoor experiments.}
\label{tab:closure}
\setlength{\tabcolsep}{3.0pt}
\begin{tabular}{c*{4}{cc}}
\toprule
\multirow{2}{*}{Data} & \multicolumn{2}{c}{EKF} & \multicolumn{2}{c}{IEKF} & \multicolumn{2}{c}{TFG-IEKF} & \multicolumn{2}{c}{TG-EqF}\\
\cmidrule(lr){2-3}\cmidrule(lr){4-5}\cmidrule(lr){6-7}\cmidrule(lr){8-9}
& m & $\%D$ & m & $\%D$ & m & $\%D$ & m & $\%D$\\
\midrule
1 & 0.047 & 0.023 & 0.071 & 0.035 & 0.072 & 0.035 & 0.070 & 0.034\\
2 & 0.190 & 0.065 & 0.205 & 0.070 & 0.215 & 0.073 & 0.241 & 0.082\\
3 & 0.173 & 0.060 & 0.097 & 0.034 & 0.096 & 0.033 & 0.097 & 0.034\\
4 & 0.077 & 0.031 & 0.083 & 0.033 & 0.078 & 0.031 & 0.088 & 0.035\\
5 & 0.052 & 0.027 & 0.031 & 0.016 & 0.033 & 0.018 & 0.031 & 0.017\\
6 & 0.051 & 0.028 & 0.062 & 0.034 & 0.063 & 0.034 & 0.062 & 0.034\\
\midrule
Mean & 0.098 & 0.039 & 0.091 & 0.037 & 0.093 & 0.037 & 0.098 & 0.039\\
\bottomrule
\end{tabular}
\end{table}

\begin{table}[t]
\centering
\caption{Statistics of the outdoor experimental datasets.}
\label{tab:dataset}
\begin{tabular}{crrrrr}
\toprule
Data & Length (m) & Speed (m/s) & Airborne (s) & Stationary (s) & ZUPTs\\
\midrule
1 & 205.50 & 1.27 & 10.52 & 4.00 & 16\\
2 & 293.72 & 1.26 &  8.46 & 2.36 & 28\\
3 & 290.23 & 1.22 &  8.95 & 2.34 & 27\\
4 & 254.31 & 1.22 &  8.68 & 2.44 & 24\\
5 & 188.89 & 0.90 &  5.97 & 2.88 & 33\\
6 & 183.52 & 1.02 &  6.16 & 2.86 & 29\\
\bottomrule
\end{tabular}
\end{table}

\begin{figure}[htbp]
  \centering
\includegraphics[width=0.6\textwidth]{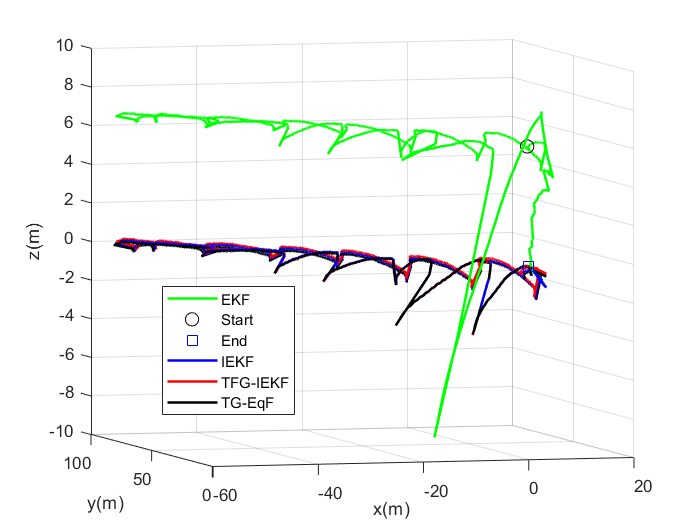}
  \caption{Trajectories of the four filters with an initial roll error of \SI{60}{\degree}.}
  \label{fig:large-roll}
\end{figure}

Absolute heading is unobservable in the present system, whereas roll and pitch can be initialized by accelerometer leveling. To examine convergence under a poor leveling solution, an initial roll error of \SI{60}{\degree} was imposed on Data 1. As shown in \cref{tab:large-roll}, the EKF accumulated approximately \SI{6}{\meter} of vertical error because of the large initial attitude error. The three geometric filters converged more rapidly and significantly reduced the vertical error. The initial vertical oscillation of the EKF is also more evident in \cref{fig:large-roll}. These results demonstrate the convergence advantage of geometric filters when the initial roll or pitch estimate is abnormal.

\begin{table}[t]
\centering
\caption{Loop-closure results for Data 1 with an initial roll error of \SI{60}{\degree}.}
\label{tab:large-roll}
\begin{tabular}{lrrrr}
\toprule
Error (m) & EKF & IEKF & TFG-IEKF & TG-EqF\\
\midrule
Horizontal position & 5.88 & 5.71 & 4.83 & 5.60\\
Height              & 6.13 & 0.48 & 0.27 & 0.63\\
\bottomrule
\end{tabular}
\end{table}

\begin{figure}[htbp!]
  \centering
\includegraphics[width=0.6\textwidth]{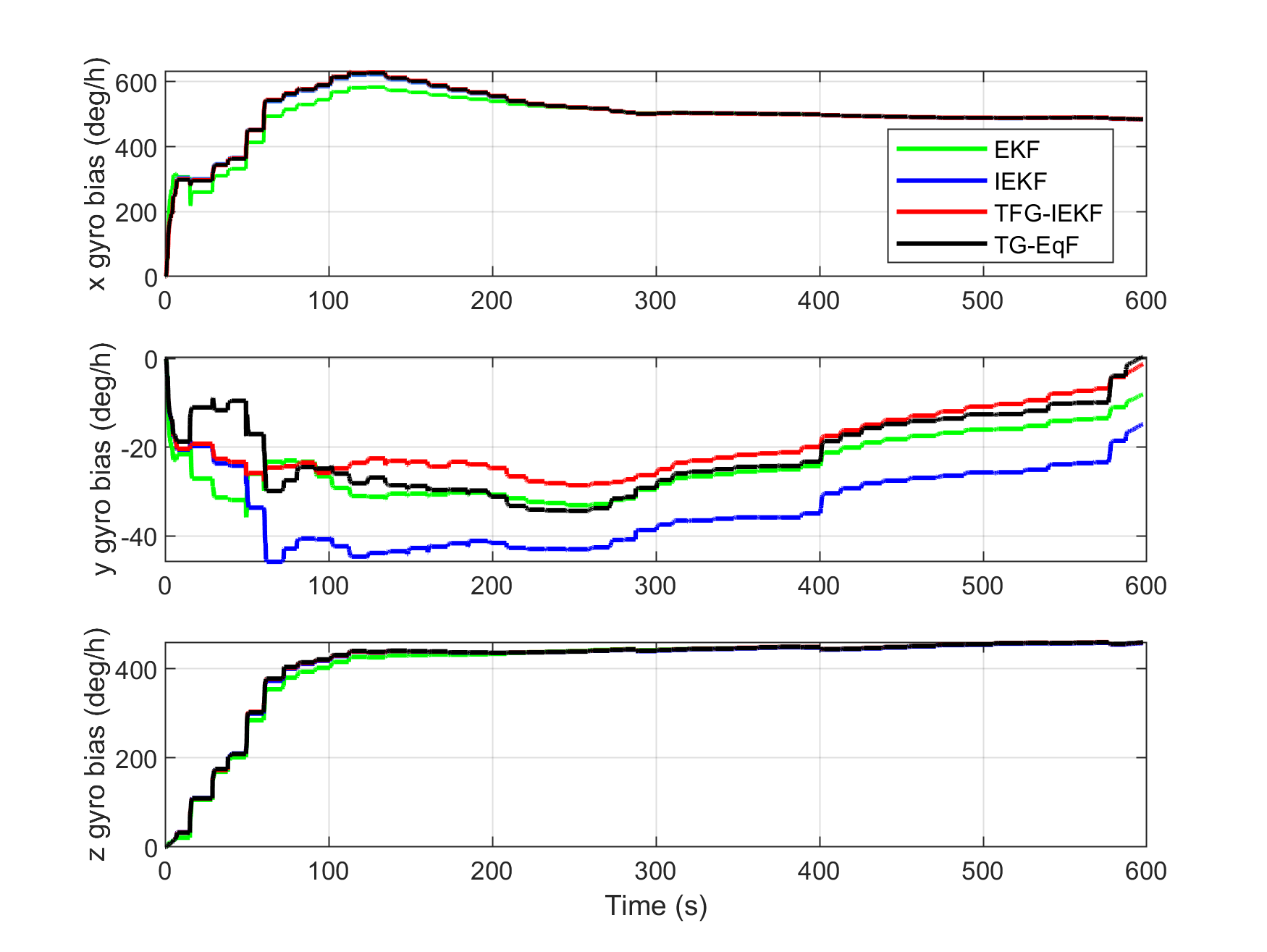}
  \caption{Gyroscope-bias estimation results.}
  \label{fig:gyro-bias}
  \vspace{-0.5cm}
\end{figure}

\begin{figure}[htbp!]
  \centering
  \includegraphics[width=0.6\textwidth]{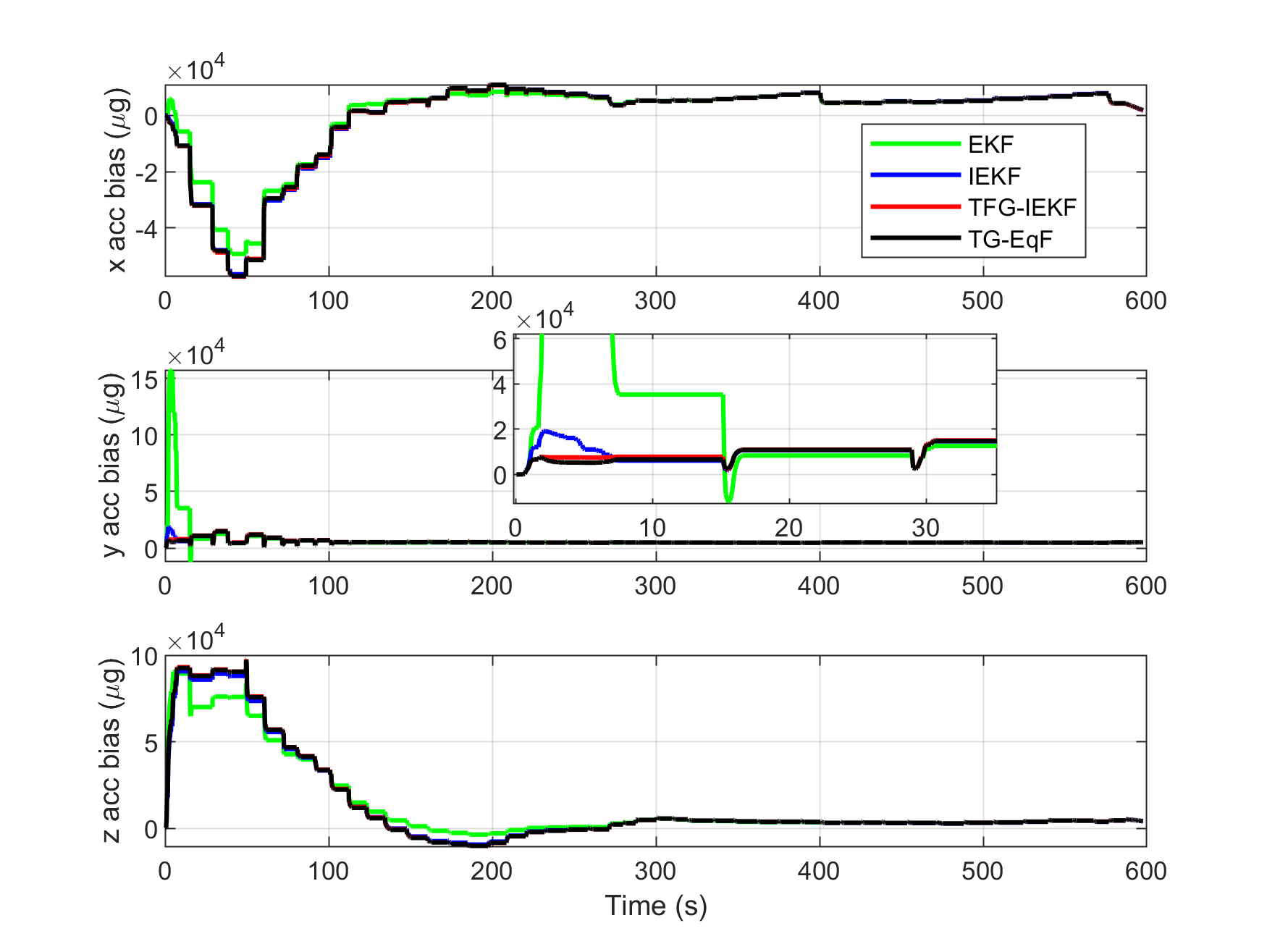}
  \caption{Accelerometer-bias estimation results.}
  \label{fig:acc-bias}
\end{figure}

Under small initial attitude errors, the four methods yielded comparable positioning accuracy and nearly identical bias estimates. To illustrate the bias convergence more clearly, a longer dataset was tested with a \SI{60}{\degree} initial roll error, and an constant gyroscope bias of \SI{500}{\degree\per\hour} and constant accelerometer bias of $5000~\mu g$ were on each axis. The bias estimates are shown in \cref{fig:gyro-bias,fig:acc-bias}. The horizontal gyroscope bias estimates followed similar convergence trends. The heading-axis bias estimates differed among different filters, but this difference has little physical significance because heading is unobservable. The three geometric filters estimated the vertical accelerometer bias substantially faster than the EKF, which contributed to better vertical position accuracy. The TFG-IEKF and TG-EqF were only marginally faster than the IEKF. As more zero-velocity observations accumulated, however, the differences among all four filters gradually disappeared.

\FloatBarrier

\subsection{Discussion}

Under normal initialization, the three geometric filters do not show a significant advantage in the INS/ZUPT system. This differs from many results reported in robotics. A principal reason is that robot systems commonly use body-frame observations, including lidar, visual-feature, and odometer measurements, which are naturally modeled using right geometric errors. Fornasier et al. \citep{fornasier2025symmetries} observed that IMU biases evolve more slowly than the navigation states. Particularly designed geometric errors can shift the navigation state-dependent linearization terms into the bias dynamics, thereby improving navigation-state convergence and consistency.

For the left-error models studied here, the comparison of \cref{eq:ekf-F,eq:ekf-G,eq:iekf-F,eq:iekf-G} shows that the IEKF removes some attitude-dependent linearization terms from the EKF system matrix. Nevertheless, $\hat{\bm{\omega}}_{ib}^b$ and $\hat{\bm{f}}_{ib}^b$ in $\bm{F}_{\mathrm{IEKF}}$ are still affected by bias-estimation error and IMU noise. Moreover, the present task is fundamentally a relative-positioning problem and the ZUPT does not provide absolute attitude information. The benefit of geometric filtering over the EKF is therefore limited under nominal conditions. As in pedestrian, visual-inertial, and lidar-inertial odometry, an estimated relative trajectory is normally aligned with a reference trajectory before evaluating the positioning accuracy. Because no continuous ground-truth trajectory was currently available in the large scale outdoor tests, loop-closure error was used here as the relative-position metric.

The Comparison of \cref{eq:iekf-F,eq:iekf-G,eq:tfg-F,eq:tfg-G,eq:tg-F,eq:tg-G} shows that the navigation-state blocks of the TFG-IEKF and TG-EqF are independent of the bias estimates. These filters can therefore be advantageous when the initial bias-estimation error is large. The bias experiments in \cref{fig:gyro-bias,fig:acc-bias}, however, show only a limited improvement over the IEKF. Absolute heading and the vertical gyroscope bias are unobservable in INS/ZUPT, and horizontal position accuracy is governed mainly by the magnitude of the vertical gyroscope bias. Consequently, when the initial roll and pitch errors are small, the four filters behave similarly. The potential advantages of the TFG-IEKF and TG-EqF might be more significant in systems with additional absolute measurements, such as tightly coupled INS/GNSS.

The EKF, IEKF, and TFG-IEKF each use a 15-dimensional error state and thus have comparable computational requirements. The TG-EqF adds a three-dimensional virtual-velocity bias, increasing its error-state dimension to 18. Because matrix multiplication and inversion dominate the filtering computational costs and have cubic complexity, the TG-EqF requires approximately $(18/15)^3\simeq1.73$ times the of computation than other filters.

\section{Conclusions}

This paper compared a conventional indirect EKF with three filters based on geometric error definitions for INS/ZUPT estimation. Unlike the right-error formulations commonly used in robotic localization and mapping, left-error TFG-IEKF and TG-EqF models were developed for reference-frame zero-velocity observations. The theoretical analysis indicates that the two models do not provide a significant consistency advantage over the left-error IEKF in this setting. Experiments with a MEMS INS/ZUPT measurement system show that, under small initial attitude errors, all three geometric filters achieve nearly the same positioning accuracy as the conventional indirect EKF. Their clearest advantage appears under an abnormally large initial roll error, where the geometric filters converge faster and substantially reduce the vertical position drift. Future work will evaluate the proposed left tangent-group equivariant formulation in INS/GNSS integrated navigation, where the requirement of absolute attitude estimation might outline stronger consistency benefits of advanced geometric filters.

\section*{Funding}

This work was supported by the National Natural Science Foundation of China under Grants 62302210, U25A20477, and 42404025.

\bibliographystyle{unsrtnat}
\bibliography{references}

@article{deng2024survey,
  author  = {Deng, Zhihong and Zhang, Ping and Li, Zhe and others},
  title   = {A Survey of Pedestrian Autonomous Navigation Technology},
  journal = {Navigation, Positioning and Timing},
  year    = {2024},
  volume  = {11},
  number  = {4},
  pages   = {1--15},
  note    = {In Chinese}
}

@article{zhao2025biped,
  author  = {Zhao, Hui and Chen, Wenbin and Liu, Ning and others},
  title   = {Inertial Positioning Method for an Individual Biped with Full-Dimensional Observable Dynamic Step Constraints},
  journal = {Journal of Chinese Inertial Technology},
  year    = {2025},
  volume  = {33},
  number  = {8},
  pages   = {743--750},
  doi     = {10.13695/j.cnki.12-1222/o3.2025.08.001},
  note    = {In Chinese}
}

@article{zhang2024vehicle,
  author  = {Zhang, Qieqie and Guo, Jingru and Lai, Jizhou},
  title   = {An Improved Zero-Velocity Detection and Motion-Constraint Method for Vehicle-Mounted Integrated Navigation},
  journal = {Journal of Chinese Inertial Technology},
  year    = {2024},
  volume  = {32},
  number  = {10},
  pages   = {1001--1009},
  note    = {In Chinese}
}

@article{lu2024interfoot,
  author  = {Lu, Yongle and Su, Sheng and Yang, Jie and others},
  title   = {Three-Dimensional Pedestrian Inertial Positioning Algorithm Assisted by Inter-Foot Distance Information},
  journal = {Navigation, Positioning and Timing},
  year    = {2024},
  volume  = {11},
  number  = {1},
  pages   = {106--114},
  note    = {In Chinese}
}

@article{zhu2022f2imu,
  author  = {Zhu, Maoran and Wu, Yuanxin and Luo, Shuang},
  title   = {{$f^2$IMU-R}: Pedestrian Navigation by Low-Cost Foot-Mounted Dual {IMUs} and Interfoot Ranging},
  journal = {IEEE Transactions on Control Systems Technology},
  year    = {2022},
  volume  = {30},
  number  = {1},
  pages   = {247--260}
}

@article{wang2026pipeline,
  author  = {Wang, Jing and Chen, Qijin and Chen, Qusen and others},
  title   = {An Incremental Observation-Correction Algorithm for {INS}/Odometer/{NHC} Fusion in Underground Pipeline Localization},
  journal = {Geomatics and Information Science of Wuhan University},
  year    = {2026},
  doi     = {10.13203/j.whugis20250194},
  note    = {In Chinese, early access}
}

@article{barrau2017stable,
  author  = {Barrau, Axel and Bonnabel, Silv{\`e}re},
  title   = {The Invariant Extended {Kalman} Filter as a Stable Observer},
  journal = {IEEE Transactions on Automatic Control},
  year    = {2017},
  volume  = {62},
  number  = {4},
  pages   = {1797--1812}
}

@article{barrau2018invariant,
  author  = {Barrau, Axel and Bonnabel, Silv{\`e}re},
  title   = {Invariant {Kalman} Filtering},
  journal = {Annual Review of Control, Robotics, and Autonomous Systems},
  year    = {2018},
  volume  = {1},
  pages   = {237--257}
}

@article{hartley2020contact,
  author  = {Hartley, Ross and Ghaffari, Maani and Eustice, Ryan M. and Grizzle, Jessy W.},
  title   = {Contact-Aided Invariant Extended {Kalman} Filtering for Robot State Estimation},
  journal = {The International Journal of Robotics Research},
  year    = {2020},
  volume  = {39},
  number  = {4},
  pages   = {402--430}
}

@article{barrau2023geometry,
  author  = {Barrau, Axel and Bonnabel, Silv{\`e}re},
  title   = {The Geometry of Navigation Problems},
  journal = {IEEE Transactions on Automatic Control},
  year    = {2023},
  volume  = {68},
  number  = {2},
  pages   = {689--704}
}

@article{goor2023eqf,
  author  = {van Goor, Pieter and Hamel, Tarek and Mahony, Robert},
  title   = {Equivariant Filter ({EqF})},
  journal = {IEEE Transactions on Automatic Control},
  year    = {2023},
  volume  = {68},
  number  = {6},
  pages   = {3501--3512}
}

@book{wang2025state,
  author    = {Wang, Maosong and Wu, Wenqi and Cui, Jiarui},
  title     = {Integrated Navigation Filtering and Estimation Based on State Transformation},
  publisher = {National Defense Industry Press},
  address   = {Beijing, China},
  year      = {2025},
  note      = {In Chinese}
}

@article{fornasier2025symmetries,
  author  = {Fornasier, Alessandro and Ge, Yixiao and van Goor, Pieter and Mahony, Robert and Weiss, Stephan},
  title   = {Equivariant Symmetries for Inertial Navigation Systems},
  journal = {Automatica},
  year    = {2025},
  volume  = {181},
  pages   = {112495}
}

@inproceedings{fornasier2022biases,
  author    = {Fornasier, Alessandro and Ng, Yimin and Mahony, Robert and Weiss, Stephan},
  title     = {Equivariant Filter Design for Inertial Navigation Systems with Input Measurement Biases},
  booktitle = {Proceedings of the 2022 IEEE International Conference on Robotics and Automation (ICRA)},
  year      = {2022},
  pages     = {4333--4339},
  address   = {Philadelphia, PA, USA}
}

@article{luo2023smoother,
  author  = {Luo, Yarong and Guo, Chi and Chen, Yichao},
  title   = {Filter and Piecewise Smoother on the Matrix {Lie} Group},
  journal = {GPS Solutions},
  year    = {2023},
  volume  = {27},
  number  = {4},
  pages   = {163}
}

@article{luo2025lefthanded,
  author  = {Luo, Yarong and Lu, Wentao and Guo, Chi and others},
  title   = {Left-Handed Symmetry Equivariant Filtering Model and Algorithm for {GNSS}/{INS} Integrated Navigation},
  journal = {Acta Geodaetica et Cartographica Sinica},
  year    = {2025},
  volume  = {54},
  number  = {8},
  pages   = {1389--1403},
  note    = {In Chinese}
}

@book{groves2013principles,
  author    = {Groves, Paul D.},
  title     = {Principles of {GNSS}, Inertial, and Multisensor Integrated Navigation Systems},
  publisher = {Artech House},
  address   = {Norwood, MA, USA},
  edition   = {2nd},
  year      = {2013}
}

@book{barfoot2017state,
  author    = {Barfoot, Timothy D.},
  title     = {State Estimation for Robotics},
  publisher = {Cambridge University Press},
  address   = {Cambridge, UK},
  year      = {2017}
}

\end{document}